\documentclass[10pt,twocolumn,letterpaper]{article}
\usepackage{iccv}
\usepackage{multirow, colortbl, graphicx, bm}
\definecolor{iccvblue}{rgb}{0.21,0.49,0.74}
\usepackage[pagebackref,breaklinks,colorlinks,allcolors=iccvblue]{hyperref}
\usepackage{comment}

\usepackage{xcolor}
\usepackage{hyperref}
\definecolor{custompink}{rgb}{0.941,0.153,0.620}
\hypersetup{
    colorlinks=true,
    linkcolor=black,
    citecolor=black,
    urlcolor=custompink  
}

\def\paperID{4628}
\def\confName{ICCV}
\def\confYear{2025}

\newcommand{\ourmethod}{PointGAC\xspace}

\newcommand{\lv}[1]{{\color{black}{#1}}}

\title{PointGAC: Geometric-Aware Codebook for Masked Point Cloud Modeling}

\author{
{\normalsize
Abiao Li$^1$, 
Chenlei Lv$^{2}$\thanks{Corresponding author.(chenleilv@mail.bnu.edu.cn)}, 
Yuming Fang$^1$,
Yifan Zuo$^1$, 
Jian Zhang$^3$,
Guofeng Mei$^4$
}\\
{\normalsize
$^1$Jiangxi University of Finance and Economics \quad
$^2$Shenzhen University
}
\\
{\normalsize
$^3$University of Technology Sydney \quad
$^4$Fondazione Bruno Kessler 
}
}

\begin{document}

\maketitle
\begin{abstract}

Most masked point cloud modeling (MPM) methods follow a regression paradigm to reconstruct the coordinate or feature of masked regions. However, they tend to over-constrain the model to learn the details of the masked region, resulting in failure to capture generalized features. To address this limitation, we propose \textbf{\textit{PointGAC}}, a novel clustering-based MPM method that aims to align the feature distribution of masked regions. Specially, it features an online codebook-guided teacher-student framework. Firstly, it presents a geometry-aware partitioning strategy to extract initial patches. Then, the teacher model updates a codebook via online k-means based on features extracted from the complete patches. This procedure facilitates codebook vectors to become cluster centers. Afterward, we assigns the unmasked features to their corresponding cluster centers, and the student model aligns the assignment for the reconstructed masked features. This strategy focuses on identifying the cluster centers to which the masked features belong, enabling the model to learn more generalized feature representations. Benefiting from a proposed codebook maintenance mechanism, codebook vectors are actively updated, which further increases the efficiency of semantic feature learning. Experiments validate the effectiveness of the proposed method on various downstream tasks. Code is available at \href{https://github.com/LAB123-tech/PointGAC}{\textcolor{custompink}{\texttt{https://github.com/LAB123-tech/PointGAC}}}.

\end{abstract} 

\vspace{-3mm}
\section{Introduction}
\label{sec:intro}
\begin{figure}[]
  \centering
  \vspace{-2mm}
  \includegraphics[width=\linewidth]{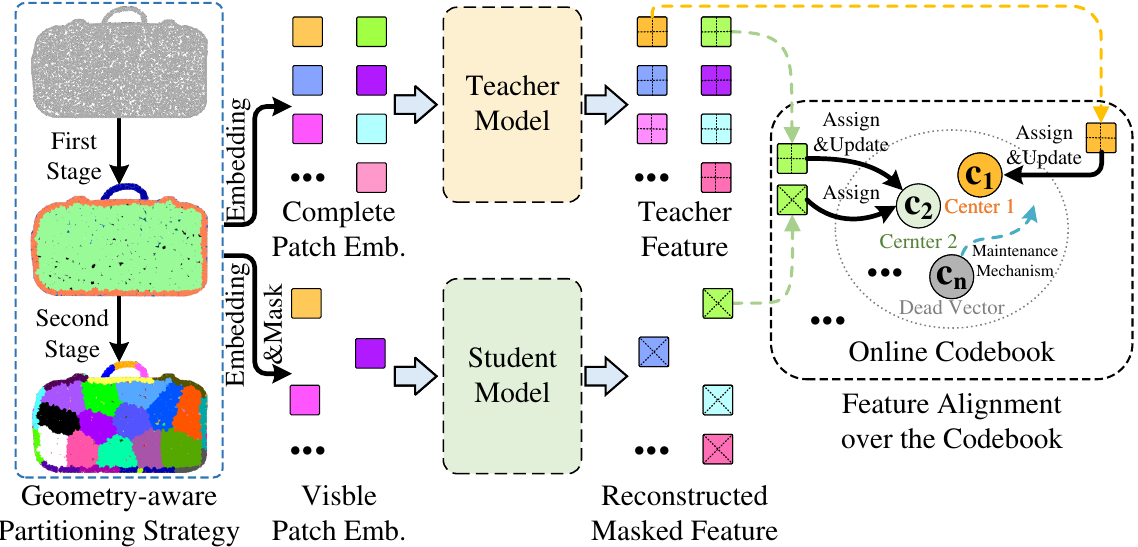}
  \caption{Fundamental concept of \textbf{\textit{PointGAC}}. Both the unmasked teacher features and the reconstructed masked features at corresponding positions are assigned to the same cluster center over the codebook. This clustering paradigm avoids forcing the student model to overfit the exact feature details of the teacher model.}
  \label{f1}
  \vspace{-6mm}
\end{figure}
Self-supervised point cloud representation learning aims to extract robust and generalizable feature representations from large-scale unlabeled data~\cite{mei2024unsupervised,mei2022data}. These learned representations can significantly improve various downstream applications, including object classification~\cite{0-3DCTN}, semantic segmentation~\cite{1-StratifiedTransformer, li2023laptran, li2025gstran}, point cloud completion~\cite{2-PoinTr}, and other 3D understanding tasks~\cite{3-CusterFormer, qu2024conditional, qu2025end}.
Due to its ability to learn without relying on annotated data, self-supervised point cloud representation learning has attracted increasing research interest. This is achieved by designing pretext tasks that encourage the model to capture meaningful geometric features. Among these, masked point cloud modeling is a prominent approach in vision transformers.

Inspired by the success of MAE~\cite{5-MAE}, masked point cloud modeling~\cite{6-PointMAE, 7-PointGame, 8-PCPMAE, 9-pointgpt, 10-PointLGMask, 11-3DOAE} aims to train models capable of reconstructing the coordinates of masked regions using the visible portions of the point cloud. However, these coordinate-based reconstruction approaches focus on capturing surface details, limiting their effectiveness in learning semantic information and resulting in inferior performance in downstream tasks.
To mitigate this limitation, recent works~\cite{12-Point2vec, 13-PosBert, 14-RIMAE, 15-3DJEPA, 16-PointMSD, 17-PointOTG,li2025cross} adopt a teacher-student framework, where instead of reconstructing coordinates, the student model is trained to align with the masked features extracted by the teacher model. \lv{However, a notable drawback of these approaches is that they tend to over-constrain the student model, forcing it to regress the point-wise feature details of the teacher, which prevents it from learning more generalized features.} Moreover, these methods often overlook geometric consistency when constructing local point cloud patches, making it challenging for the model to learn meaningful representations.

\lv{To address these issues, we propose \textbf{\textit{PointGAC}}, a novel clustering-based approach that aims to learning the similarity of features in their distribution, as illustrated in \cref{f1}}. \ourmethod begins by partitioning the input point cloud into non-overlapping, geometrically homogeneous patches. After embedding the patches, the teacher model updates an online codebook by leveraging the features extracted from the complete patch embeddings. As a result, codebook vectors gradually evolve into cluster centers. The student model is tasked with reconstructing the masked features from the visible patch embedding. Finally, the reconstructed masked features and the unmasked teacher features at the corresponding locations are assigned to the same cluster center to achieve feature alignment. This clustering paradigm enables the student model to learn the cluster centers to which the masked point cloud patches belong. \lv{It avoids the need of point-wise feature alignment, facilitating the learning of more generalized feature representations in the student network.} Moreover, we find that many vectors in the codebook remain unupdated and become dead vectors. Therefore, we design a codebook maintenance mechanism to revitalize these dead vectors. It helps to improve the efficiency of masked feature alignment over the codebook. Our method presents a general solution for point cloud-based semantic analysis. Experimental results demonstrate its efficacy across a variety of downstream tasks. The contribution can be summarized as follows:

\begin{itemize}    

    \item We propose a novel online codebook-guided teacher-student framework that follows a clustering-based paradigm to accomplish MPM tasks. This framework predicts the cluster center to which the masked region belongs, effectively avoiding the problem of over-constraining the model.
    
    \item We propose a geometry-aware partitioning strategy to divide the point cloud into geometrically homogeneous patches. This strategy ensures geometric consistency within each local point cloud patch, enhancing the quality and efficiency of feature learning.
    
    \item We design a codebook maintenance mechanism to increase the update frequency of the codebook vectors. It improves the efficiency of masked feature alignment.
    
\end{itemize}

\section{Related Work}

\noindent\textbf{\textit{Masked point modeling}.} Inspired by the success of Mask Image Modeling (MIM), MPM has also gained increasing attention in point cloud research. The pioneering algorithm \cite{18-PointBert} introduces a trained tokenizer, allowing pretraining to be performed through masked token prediction. However, the tokenizer is trained offline, making it immutable during pretraining. Other works\cite{10-PointLGMask, 20-MAE3D, 21-PointFEMAE} explore multi-task learning to embed both local and global contexts for point cloud pretraining. Several studies \cite{19-PointM2AE, 7-PointGame, 17-PointOTG} are dedicated to designing complex models, leading to significant advancements in the reconstruction of point cloud geometric structures. However, the model spends significant resources to accurately recover the position of each point, which may not be important for downstream tasks related to semantic understanding. Hence, certain algorithms \cite{12-Point2vec, 13-PosBert, 17-PointOTG, 15-3DJEPA, 16-PointMSD} shift their focus toward reconstructing the features of masked regions.

Although existing methods have achieved impressive results, they are regression-based approaches for reconstructing the information of the masked regions. Such regression-based algorithms tend to over-constrain the model, leading it to focus excessively on reconstructing details while failing to learn more generalizable features. In this paper, we adopt a clustering-based method to identify which cluster center in the codebook the masked features belong to. This allows the model to avoid this problem.


\noindent\textbf{\textit{Codebook for representation learning}.} The codebook is widely used as a feature representation method in computer vision. In generative models, \cite{25-VQVAE, 26-DVAE} leverages the codebook to map continuous features to discrete tokens, facilitating the diverse visual outputs from the textual descriptions. Inspired by this, subsequent studies~\cite{27-BEIT, 28-HumanPose} represent the object as a series of discrete tokens to capture structural patterns. Several methods~\cite{29-SWAV, 30-OBOW, 31-MaskSiamese, 51-PAWS} utilize a codebook to align representations of different augmented views in the clustering space. SwAV~\cite{29-SWAV} adopts an optimal transport strategy~\cite{37-Sinkhorn} that assumes a uniform assignment of each batch data to all prototypes, while MSN~\cite{31-MaskSiamese} and PAWS~\cite{51-PAWS} employ entropy-based regularization~\cite{53-MeanRegulation} to encourage balanced prototype usage. However, such an assumption may not hold in our setting due to the imbalanced and potentially diverse semantic distribution within each mini-batch. Instead, we adopt an online k-means algorithm to progressively update the codebook and introduce a maintenance mechanism to improve its utilization efficiency.

\section{Method}
\begin{figure*}[t]
  \centering
\vspace{-2mm}
  \includegraphics[width=\linewidth]{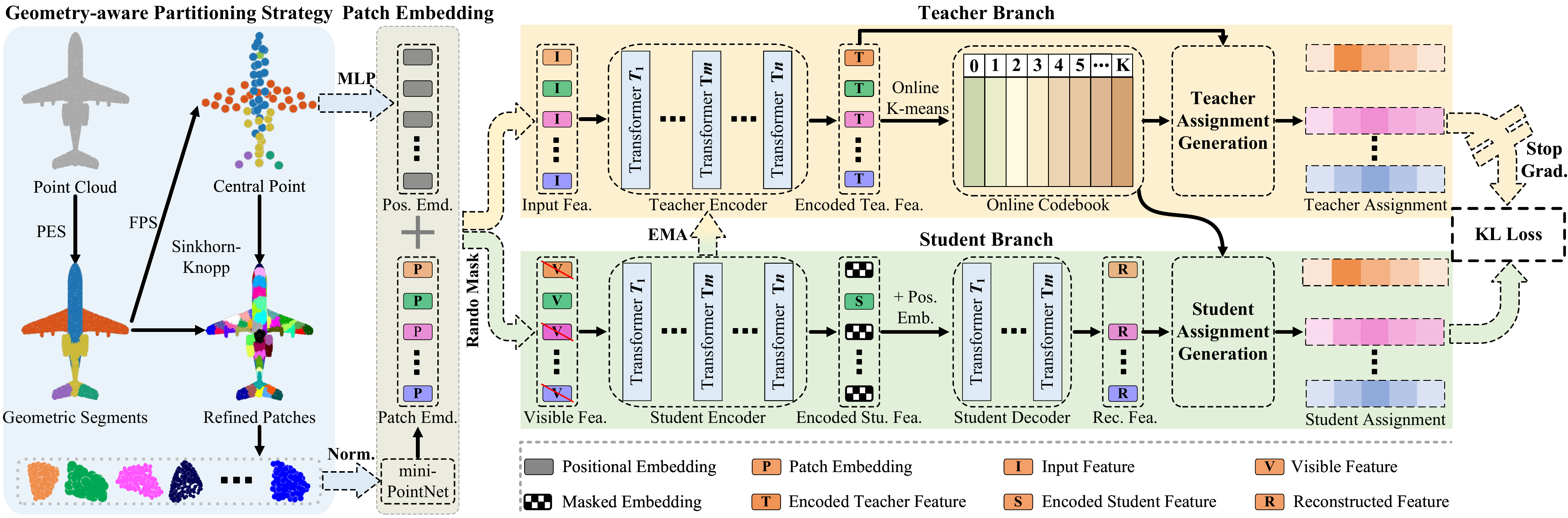}
  \caption{The framework of \textbf{\textit{PointGAC}}. This framework utilizes a online codebook to achieve feature alignment. It aligns the reconstructed features on the student side with the encoded features on the teacher side at corresponding positions.}
  \label{f2}
  \vspace{-0.5cm}
\end{figure*}

\textbf{\textit{Overview.}} \cref{f2} shows the architecture of the proposed \ourmethod, adopting an online codebook-guided teacher-student framework.  Given an input point cloud \textit{P} with $N$ points, we first partition it into $L$ non-overlapping refined patches using a geometry-aware partitioning strategy. The central points are then processed through an MLP layer to obtain the positional embedding for each patch. Each patch undergoes the operation of normalization before being passed into a mini-PointNet for patch embedding. All patch and positional embeddings are summed to form input features for subsequent teacher-student framework.

In the teacher branch, input features are fed into an encoder to update codebook vectors via online k-means clustering. Encoded teacher features are matched with the codebook to generate teacher assignments, which serve as the target for the student model to learn. In the student branch, the task is to recover the masked features from partially visible features, obtaining the reconstructed features. Then, the reconstructed features are accomplished by matching with the codebook to generate student assignments. Then, the teacher and student assignments are aligned using a KL divergence loss. During training, gradients are propagated only through the student branch, while the teacher encoder is updated using an Exponential Moving Average (EMA) of the student encoder's parameters. In the following, we will detail the key components of the proposed algorithm.

\subsection{Geometry-aware Partitioning Strategy}
In MIM tasks, due to the regular and structured nature of images, algorithms~\cite{31-MAE, 32-BEIT, 33-IBOT} typically divide images into non-overlapping grids. However, since point clouds are unordered and unstructured data, the vast majority of current algorithms rely on a combination of FPS (Farthest Point Sampling) and KNN (K-Nearest Neighbors) to construct overlapping point cloud patches. A major drawback of these methods is the leakage of information from visible patches to masked patches due to the information redundancy between patches, which reduces the practicality. To address this issue, Wang~{\em et al.}~\cite{17-PointOTG} introduced an optimal transport-based approach to partition the point cloud into non-overlapping patches. However, it overlooks the geometric properties of the point cloud.

To overcome such limitations, we propose a geometry-aware partitioning strategy that divides a point cloud into non-overlapping and geometrically consistent patches, as shown in~\cref{f2} (blue area). To group points with similar geometric features, we apply Potts Energy Segmentation (PES)~\cite{34-LargeCut, 35-GlbalEnergy} for coarse partitioning, formulated as:
\begin{equation}
\scalebox{1.0}{%
$\displaystyle
\arg\min_{g{\in}\mathbb{R}^{dg}}\sum_{i\in P}\left\|g_i{-}f_i\right\|^2{+}\mu\sum_{(i,j){\in} E_{\mathrm{nn}}}w_{ij}\left[g_i{-}g_j{\neq} 0\right]
$%
},
\label{e1}
\end{equation}
where $f_i$ is the original geometric feature of point $i$, as defined in \cite{34-LargeCut}, $g_i$ is the target geometric label, $P$ represents the point cloud, $E_{nn}$ is the set of edges that connect neighboring points, defining the local neighborhood structure, $w_{ij}$ is the weight determining the influence of the pairwise interaction between points $i$ and $j$. The Iverson bracket $[\cdot]$ equals 1 if the condition holds and 0 otherwise. The factor $\mu>0$ regulates the coarseness of the resulting partition. This enables the generation of meaningful, non-overlapping geometric segments.

In MPM, a fixed number of patches is required, whereas PES produces a varying number of segments across different point clouds.
To reconcile these differences, we use FPS to sample $L$ central points (along with their geometric labels) from the geometric segments. Then, we assign each point to related central point, thereby forming $L$ refined patches. This assignment satisfies two conditions: 1) All points within a patch belong to the same geometric segment; 2) Different patches belong to the same segment contain a similar number of points. We denote the assignment by a matrix $Q$, whose elements $Q_{ij}\in [0,1]$.

To enforce the first condition, we define a mask matrix $M$ with elements $M_{ij}\in \{0,1\}$, where $M_{i j}=1$ if the central point and its assigned points share the same geometric label, and 0 otherwise.  We then apply $M$ to $Q$ via the Hadamard (element-wise) product $Q \circ M$ to guide the assignments. The second condition can be approximated as $\sum_i (Q \circ M)_{ij} / L{=} {N}/{L}$. Since FPS sampling tends to select more cenral points from segments containing a larger number of points, it naturally balances the partitioning. Consequently, there is no need to enforce explicit constraints at the level of each individual segment. To satisfy both constraints, we formulate the assignment process as a \emph{masked optimal transport}, expressed as:
\begin{equation}
    \min _{Q} \sum_{i=1}^{N} \sum_{j=1}^{L} M_{i j}Q_{i j} T_{i j}, \\
    \text{s.t.,} \sum_{i=1}^N \frac{(Q \circ M)_{ij}}{L} = \frac{N}{L},
\end{equation}
where $T$ is the cost. Following \cite{36-MOT}, the solution to this problem is achieved based on Sinkhorn algorithm~\cite{37-Sinkhorn}.
This process yields refined patches $\{P_1, \cdots, P_L\}$ that are geometrically aware, ensuring each patch consists exclusively of points from the same geometric segment.

\noindent\textbf{\textit{Patch embedding layer.}} To extract features from the achieved patches, patch embedding layers are designed. Each refined patch is normalized by subtracting its center. The resulting patches are then processed by a mini-PointNet~\cite{38-PointNet} to generate patch embeddings $E \in \mathbb R^{L\times D}$ ($D$ is the dimension), which can be expressed as:
\begin{equation}
    E = \text{PointNet}\left(\{P_1, \cdots, P_L\}\right),
\end{equation}
Notably, patch sizes are not uniform. To enable parallel feature extraction across different patches, we construct a mask matrix $H \in \mathbb R^{N\times N}$ over the entire point cloud, where $H_{i,j}=1$ indicates that points $i$ and $j$ belong to the same patch. 
Max pooling is then applied to regions where the mask matrix has a value of 1 in each row. Meanwhile, position embeddings $PE \in \mathbb R^{L\times D}$ are extracted from the patch center using an MLP layer. Finally, the patch embeddings $E$ and the positional embeddings $PE$ are summed to obtain $F$, which serves as input features for subsequent modules.

\subsection{Teacher-Student Framework}  
To achieve feature alignment in a self-supervised manner, we employ a teacher-student framework, where the student branch reconstructs masked features based on the knowledge provided by the teacher branch.

\noindent\textbf{\textit{Teacher branch}}. Given the input feature \textit{F}, the encoder applies a self-attention mechanism to encode \textit{F}. It is composed of \textit{n} standard transformer blocks. Specially, the encoded teacher feature within the \textit{i}-th transformer block of the encoder can be formulated as:
\begin{equation}
T_{i}=\operatorname{Attn}\left(Q, K, V\right)=\operatorname{SoftMax}\left(\frac{Q K}{\sqrt{D}}\right)V,
\end{equation}
where $Q = T_{i-1}W_Q$, $K = T_{i-1}W_K$, and $V = T_{i-1}W_V$, with  $W_Q$, $W_K$, and $W_V$ being learnable parameters. Furthermore, $T_0 = F$.

Subsequently, we utilize the encoded teacher features to update the codebook $C = [c_1, c_2, \dots, c_K]$  where each vector $c_k \in \mathbb R^{1\times D}$ represents a D-dimensional vector.  The codebook is updated using an online k-means algorithm~\cite{25-VQVAE}. For each teacher feature, we identify the most similar codebook vector. We use $n_k^{(t)}$ to denote the number of features assigned to $c_k$ at the $t$-th iteration in the pre-training. We also use $m_k^{(t)}$ to represent the element-wise sum of these features. To maintain a stable update processing, we utilize an exponential moving average:
\begin{equation}
\vspace{-2mm}
    \begin{aligned}
        N_k^{(t)}=\gamma N_k^{(t-1)}+(1-\gamma)n_k^{(t)},\\
        M_k^{(t)}=\gamma M_k^{(t-1)}+(1-\gamma)m_k^{(t)},\\
    \end{aligned}
\end{equation}
where $N_k^{(t)}$ is the accumulated count of the features assigned to $c_k$ at the $t$-th iteration, with $N_k^{(1)}=1$ as the initial value. $M_k^{(t)}$ is the accumulated element-wise sum of these features and $M_k^{(1)}$ is initialized as the $k$-th codebook vector. Then, $c_k$ is given by $c_k=M_k/N_k$. We set $\gamma = 0.99$ to ensure that the update processing remains stable, allowing the codebook vector to become the cluster center gradually.

\noindent\textbf{\textit{Teacher assignment generation}}. To assign the encoded teacher features to cluster centers within the codebook, we compute the soft assignment of encoded teacher features across the codebook. Given the encoded teacher features $T = \{t_i\}_{i=1}^L, t_i \in \mathbb R^{1\times D}$, the soft assignment $Q_t\in \mathbb R^{L\times K}$ 
is computed as the following formula:
\begin{equation}
    Q_t=\mathrm{SoftMax}\left(\frac{TC}{\tau_{t}}\right),
\end{equation}
where $\tau_t$ is the temperature coefficient used to adjust the smoothness of $Q_t$. In our experiments, we adopt a cosine annealing strategy to adjust the value of $\tau_t$. At the early stage, $\tau_t$ is relatively large, resulting in smoother distribution. As training progresses, we gradually reduce $\tau_t$ , which results in highly peaky distribution. Essentially, $Q_t$ encodes the cluster center over the codebook to which the teacher feature belongs. We use these generated assignment vectors $Q_t$ as the supervision signals for the student branch.

\noindent\textbf{\textit{Student branch}.} The task of the student branch is to reconstruct the masked feature from the visible features. With a predefined masking ratio $r$, we apply random masking to the input feature $F$. The masked features are denoted as $S_m \in \mathbb R^{\lfloor Lr \rfloor \times D}$, and the visible features are denoted as $S_v \in \mathbb R^{\lceil L(1-r) \rceil \times D}$, where $\lfloor \cdot \rfloor$  and $\lceil \cdot \rceil $denote the floor and ceiling functions, respectively. On the student side, apart from possessing an encoder with the same structure as the teacher model, it also includes a decoder that comprises \textit{m} transformer modules. The student encoder extracts features from $S_v$, yielding the encoded student features $S_E$. Then, $S_E$ is concatenated with the learnable mask embeddings $S_a \in \mathbb{R}^{\lfloor Lr \rfloor \times D}$ and added to the full set of positional information $PE$, serving as the input to the decoder. After being processed by the decoder, the reconstructed feature $S_R \in \mathbb{R}^{Lm \times D}$ is used to align the encoded teacher features. 

\noindent\textbf{\textit{Student assignment generation}.} The reconstructed features $S_R$ and the teacher features $T$ at the masked locations should be assigned to the same cluster center to achieve feature alignment. We compute the soft assignment $Q_s\in \mathbb R^{L\times K}$ of the features $S_R$ over the codebook $C$ using the following formula:
\begin{equation}
\vspace{-2mm}
    Q_s=\mathrm{SoftMax}\left(\frac{S_DC}{\tau_{s}}\right),
\end{equation}
where $\tau_t$ is the temperature parameter that controls the sharpness of the soft assignment. In the experiments, we set $\tau_s$ to a constant value that is larger than $\tau_t$. It ensures the student assignment $Q_s$ is smoother in distribution compared to the teacher assignment $Q_T$, helping the reconstructed feature to better align to the teacher feature.
\subsection{Codebook Maintenance Mechanism}
\begin{figure}[ht]
  \centering
  \includegraphics[width=\linewidth]{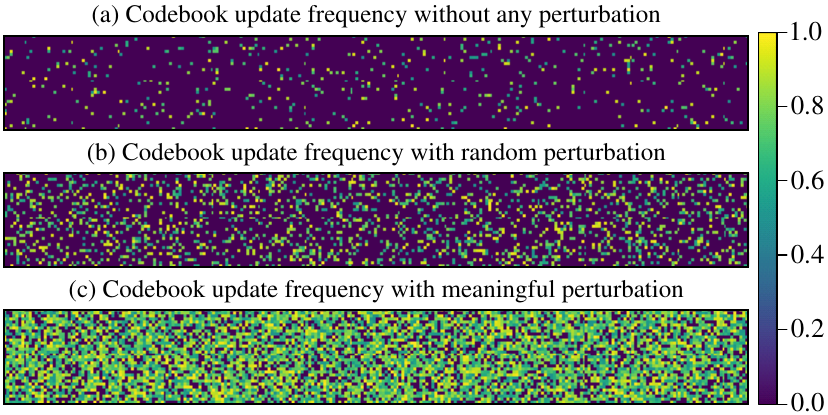}
  \caption{Visualization comparison of codebook update frequency under different perturbation modes. Each pixel in the figure represents the update frequency of a single vector. }
  \label{f3}
  \vspace{-0.5cm}
\end{figure}

The codebook vectors are designed to serve as cluster centers for the patch features. However, many vectors remain unused and gradually become 'dead vectors' since online k-means are based on the nearest-neighbor lookup. To analyze this issue, we record the update frequency of all codebook vectors after training. The frequency matrix (1, K) was reshaped to (H, W) for visualization. As shown in Fig.~\ref{f3}(a), without any perturbation, many codebook vectors exhibit low update frequencies. For this issue, we experimentally introduced random perturbations to codebook vectors during training and observed it improved codebook utilization to some extent, as shown in \cref{f3}(b).

Building on this insight, we propose a codebook maintenance mechanism, which applies varying degrees of meaningful perturbations to each codebook vector based on its update frequency. For each codebook vector $c_k$, we assign the most similar teacher feature $t_k$ from the current batch of data to update it. The principle is that frequently updated codebook vectors receive minimal perturbations, whereas infrequently updated or dead vectors undergo larger perturbations. Specially, we count the number of times the codebook vector is updated, denoted as $x$. Then, we compute a weight that determines the magnitude of the perturbation to be added.  The weights $\alpha_k$ are computed as:
\begin{equation}
   \begin{aligned}
       \alpha_k &= \frac{1}{1+e^{\epsilon(x-\bar x)}}, \\
        c_k &\leftarrow c_k\cdot(1-\alpha_k)+t_{k}\cdot\alpha_k,
   \end{aligned}
\end{equation}
where $\bar{x}$ is the average of the maximum and minimum update counts among all the codebook vectors. $\epsilon$ is a small value used to smoothly adjust the weights assigned to different codebook vectors. As illustrated in \cref{f3}(c), the majority of the codebook vectors exhibit a high update frequency, indicating that they are being sufficiently updated at this stage. Experiments discuss the improvement.

\section{Experiments}
\noindent\textbf{\textit{Pre-training dataset}.} We pre-trained the proposed model on ShapeNet55~\cite{39-ShapeNet}, which consists of 52,472 unique 3D models covering 55 object categories. Before pre-training, we apply Potts energy segmentation on the point cloud to complete the geometric partitioning. Geometric labels are stored in the dataset to speed up pre-training. 

\noindent \textbf{\textit{Experimental details}.} We sample $N = 1024$ points with their corresponding geometric labels from the dataset using FPS and partition them into $L=64$ patches. We employ the AdamW optimizer \cite{40-adam} with a weight decay of 0.04 for training. The learning rate follows a cosine decay schedule \cite{41-SGDR}, starting with a 50-epoch linear warm-up phase and reaching a maximum learning rate of 0.001. $\tau_t$ is initialized at 0.07 and decay to 0.04 during training. $\tau_s$ remains fixed at 0.1 throughout the entire training process. Following the practice in OBoW~\cite{30-OBOW}, the codebook size $K$ is set to 8192.

\noindent \textbf{\textit{Model architecture}.} The teacher and student encoders share the same structure, both composed of $n=12$ transformer blocks. The student decoder consists of $m=4$ standard transformer blocks. The hidden dimension of transformer blocks is $D=384$.

\subsection{Evaluations on Downstream Tasks}
\begin{table*}[ht]
\centering
\small
\tabcolsep 1.5pt
{%
\begin{tabular}{lccccccccccc}
\toprule
\multirow{2}{*}{Method} & \multirow{2}{*}{References} & \multirow{2}{*}{\#Params(M)} & \multicolumn{4}{c}{ModelNet40} & \multicolumn{4}{c}{ScanObjectNN} \\ 
\cmidrule(lr){4-7}  \cmidrule(lr){8-11}
 &  &  & \#Points & Linear SVM & w/o Vote & w/ Vote & \#Points & OBJ-BG & OBJ-ONLY & PB\_T50\_RS \\ 
\midrule
Point-BERT\cite{18-PointBert} & CVPR 2022 & 22.1 & 1k points & 87.4 & 92.7 & 93.2 & 1k points & 87.4 & 88.1 & 83.1 \\
PointMAE\cite{6-PointMAE} & ECCV 2022 & 22.1 & 1k points & 92.7 & 93.2 & 93.8 & 2k points & 90.0 & 88.3 & 85.2 \\
PointM2AE\cite{19-PointM2AE} & NeurIPS 2022 & 15.3 & 1k points & 92.9 & 93.4 & 94.0 & 2k points & 91.2 & 88.8 & 86.4 \\
Point2Vec\cite{12-Point2vec} & GCPR 2023 & 23.2 & 1k points & 92.1 & 92.5 & 93.1 & 2k points & 91.2 & 90.4 & 87.5 \\
PointGPT\cite{9-pointgpt} & NeurIPS 2023 & 19.5 & 1k points & - & - & 94.0 & 2k points & 91.6 & 90.0 & 86.9 \\
MAE3D\cite{20-MAE3D} & TMM 2023 & 23.4 & 1k points & - & 93.4 & - & 2k points & 88.4 & 87.7 & 86.2 \\
PointGame\cite{7-PointGame} & TGRS 2023 & - & 1k points & 90.0 & 93.1 & 93.5 & 2k points & 89.0 & 89.0 & 85.8 \\
MaskFeat3D\cite{42-MaskFeat3D} & ICLR 2024 & - & 1k points & 91.1 & 93.3 & 93.7 & 2k points & 91.7 & 90.0 & 87.7 \\
RIMAE\cite{14-RIMAE} & AAAI 2025 & 23.8 & 1k points & 92.8 & 93.1 & 93.9  & 2k points &  91.9 & 89.3 &   86.8              \\
\midrule
\rowcolor{gray!20} Ours & - & 24.1 & 1k points & \textbf{93.5} & \textbf{94.3} & \textbf{94.6} & 2k points & \textbf{92.6} & \textbf{91.4} & \textbf{89.4} \\
\bottomrule
\end{tabular}%
}
\vspace{-0.3cm}
\caption{Classification accuracy on ModelNet40 and ScanObjectNN. \#Params indicates model size. For ModelNet40, overall accuracy (\%) is reported with and without voting. For ScanObjectNN, overall accuracy (\%) is shown for three variants.}
\label{t1}
\end{table*}

\begin{table}[t]
\centering
\small
\resizebox{\columnwidth}{!}
{
\begin{tabular}{lcccc}
\toprule
\multicolumn{1}{l}{\multirow{2}{*}{Method}} & \multicolumn{2}{c}{5-way} & \multicolumn{2}{c}{10-way} \\
\cmidrule(lr){2-3} \cmidrule(lr){4-5}
\multicolumn{1}{c}{} & 10-shot & 20-shot & 10-shot & 20-shot \\
\midrule
Point-BERT\cite{18-PointBert} & 94.6\textnormal{\scriptsize{$\pm$3.6}} & 93.9\textnormal{\scriptsize{$\pm$3.1}} & 86.4\textnormal{\scriptsize{$\pm$5.4}} & 91.3\textnormal{\scriptsize{$\pm$4.6}} \\
MaskPoint\cite{54-MaskPoint} & 95.0\textnormal{\scriptsize{$\pm$3.7}} & 97.2\textnormal{\scriptsize{$\pm$1.7}} & 91.4\textnormal{\scriptsize{$\pm$4.0}} & 92.7\textnormal{\scriptsize{$\pm$5.1}} \\
Point-MAE\cite{6-PointMAE} & 96.3\textnormal{\scriptsize{$\pm$2.5}} & 97.8\textnormal{\scriptsize{$\pm$1.8}} & 92.6\textnormal{\scriptsize{$\pm$4.1}} & 93.4\textnormal{\scriptsize{$\pm$3.5}} \\
Point-M2AE\cite{19-PointM2AE} & 96.8\textnormal{\scriptsize{$\pm$1.8}} & 98.3\textnormal{\scriptsize{$\pm$1.4}} & 92.3\textnormal{\scriptsize{$\pm$4.5}} & 95.0\textnormal{\scriptsize{$\pm$3.0}} \\
PointGame\cite{7-PointGame} & 96.7\textnormal{\scriptsize{$\pm$2.7}} & 97.9\textnormal{\scriptsize{$\pm$1.5}} & 92.9\textnormal{\scriptsize{$\pm$3.7}} & 94.7\textnormal{\scriptsize{$\pm$2.6}} \\
MAE3D\cite{20-MAE3D} & 95.2\textnormal{\scriptsize{$\pm$3.1}} & 97.9\textnormal{\scriptsize{$\pm$1.6}} & 91.1\textnormal{\scriptsize{$\pm$4.6}} & 95.3\textnormal{\scriptsize{$\pm$3.1}} \\
PointOTG\cite{17-PointOTG} & 97.2\textnormal{\scriptsize{$\pm$2.3}} & 98.7\textnormal{\scriptsize{$\pm$1.2}} & 93.2\textnormal{\scriptsize{$\pm$3.4}} & 95.6\textnormal{\scriptsize{$\pm$2.6}} \\
\midrule
\rowcolor{gray!20} Ours & \textbf{97.4}\textnormal{\scriptsize{\bm{$\pm$}\textbf{2.0}}} & \textbf{98.9}\textnormal{\scriptsize{\bm{$\pm$}\textbf{1.3}}} & \textbf{93.7}\textnormal{\scriptsize{\bm{$\pm$}\textbf{3.1}}} & \textbf{95.9}\textnormal{\scriptsize{\bm{$\pm$}{2.4}}} \\
\bottomrule
\end{tabular}
}
\vspace{-0.3cm}
\caption{Few-shot classification results on ModelNet40. The mean accuracy and standard deviation are calculated in each setting.}
\label{t2}
\end{table}
\begin{table}[t]
\centering
\small
\tabcolsep 1.5pt
{
\begin{tabular}{lcccc}
\toprule
\multicolumn{1}{l}{\multirow{2}{*}{Method}} & \multicolumn{2}{c}{Semantic segmentation} & \multicolumn{2}{c}{Part segmentation} \\
\cmidrule(lr){2-3} \cmidrule(lr){4-5}
\multicolumn{1}{c}{} & ~~~~mAcc & ~~~mIoU & mIoU$_C$ & mIoU$_I$ \\
\midrule
Transformer\cite{43-Transformer} & ~~~~68.6 & ~~~60.0 & 83.4 & 84.7 \\
Point-MAE\cite{6-PointMAE} & ~~~~69.9 & ~~~60.8 & 84.2 & 86.1 \\
Point-BERT\cite{18-PointBert} & ~~~~69.7 & ~~~60.5 & 84.1 & 85.6 \\
PCP-MAE\cite{8-PCPMAE} & ~~~~70.4 & ~~~61.1 & 84.9 & 86.1 \\
Point-LGMask\cite{10-PointLGMask} & ~~~~70.3 & ~~~61.3 & 84.4 & 86.1 \\
PointMPP\cite{45-PointMPP} & ~~~~69.8 & ~~~61.4 & 84.7 & 86.1 \\
RIMAE\cite{14-RIMAE} & ~~~~68.4 & ~~~60.3 & 82.1 & 84.3 \\
\midrule
\rowcolor{gray!20} Ours & ~~~~\textbf{70.9} & ~~\textbf{61.8} & \textbf{85.1} & \textbf{86.7} \\
\bottomrule
\end{tabular}%
}
\vspace{-0.3cm}
\caption{Segmentation results on S3DIS and ShapeNetPart. For S3DIS, mean instance IoU (mIoU) and mean class accuracy (mAcc) are reported; for ShapeNetPart, category-level (mIoU$_C$) and instance-level (mIoU$_I$) mean IoU are provided.}
\vspace{-0.4cm}
\label{t3}
\end{table}

\begin{table}[t]
\centering
\small
\tabcolsep 5pt
{
\begin{tabular}{lccccc}
\toprule
Method & CD-S & CD-M & CD-H & CD-Avg & F1 \\
\midrule
Point-JEPA\cite{48-Pointjepa} & 0.83 & 1.23 & 2.06 & 1.37 & 0.34 \\
Pos-Bert\cite{13-PosBert} & 0.91 & 1.27 & 2.25 & 1.48 & 0.32 \\
3D-JEPA\cite{15-3DJEPA} & 0.71 & 1.22 & 2.02 & 1.32 & 0.35 \\
Point2Vec\cite{12-Point2vec}  & 0.58 & 0.81 & 1.92 & 1.10 & 0.38 \\
Point-MSD\cite{16-PointMSD} & 0.65 & 0.83 & 1.87 & 1.12 & 0.37 \\
PM-MAE\cite{49-PMMAE} & 0.53 & 0.79 & 1.81 & 1.04 & 0.41 \\
RIMAE\cite{14-RIMAE} & \textbf{0.51} & 0.72 & 1.74 & 0.99 & 0.46 \\
\midrule
\rowcolor{gray!20} Ours & 0.52 & \textbf{0.69} & \textbf{1.63} & \textbf{0.95} & \textbf{0.49} \\
\bottomrule
\end{tabular}%
}
\caption{Results of point cloud completion on ShapeNet55 dataset. We use CD-S, CD-M, and CD-H to denote CD results in the simple, moderate, and hard settings, respectively.}
\vspace{-0.5cm}
\label{t4}
\end{table}

\textbf{\textit{Shape classification.}} We evaluate our method by freezing the pre-trained student encoder and training a linear SVM on ModelNet40 for shape classification. We also report results with and without the voting trick after fine-tuning on ModelNet40. As shown in \cref{t1}, compared to other teacher-student models following the regression paradigm, such as Point2Vec~\cite{12-Point2vec} and RIMAE~\cite{14-RIMAE}, our method achieves the best performance in the Linear SVM evaluation. Since regressing the teacher features requires an accurate matching of each feature value, which makes the optimization process more challenging. By aligning the cluster centers to which the features belong, we only need to focus on the similarity of the feature distributions, thus simplifying the optimization process.

We further validate our model on the ScanObjectNN dataset, which contains around 15 000 real-world objects from 15 categories. We report results on three subsets: OBJ-BG (objects with background), OBJ-ONLY (objects only), and PB-T50-RS (objects with background and perturbations). To ensure fairness, each input point cloud is sampled to 2048 points. As shown in \cref{t1}, \ourmethod achieves competitive performance, particularly in challenging settings.  Although the algorithm is pre-trained on the synthetic dataset, it generalizes well to real-world datasets and achieves the best performance. Also, on the OBJ\_ONLY variant, our method shows significant performance improvement. Since the results of geometric partitioning remain unaffected by background noise. This property makes it advantageous for codebook vectors to serve as geometrically well-defined cluster centers.

\noindent \textbf{\textit{Few-shot classification}.} We perform few-shot experiments on the ModelNet40 dataset using the $m$-way, $n$-shot setting, where $m \in \{5,10\}$ denotes the number of randomly selected categories and $n \in \{10,20\}$ indicating the number of samples per category. During testing, we randomly selected 20 unseen samples from each category. In experiments, we performed 10 independent trials for each setting. The experimental results are shown in \cref{t2}. It can be observed that even with limited data, our method exhibits competitive performance compared to other methods, while showing minimal deviation in most cases.

\begin{figure*}[t]
  \centering
  \includegraphics[width=\linewidth]{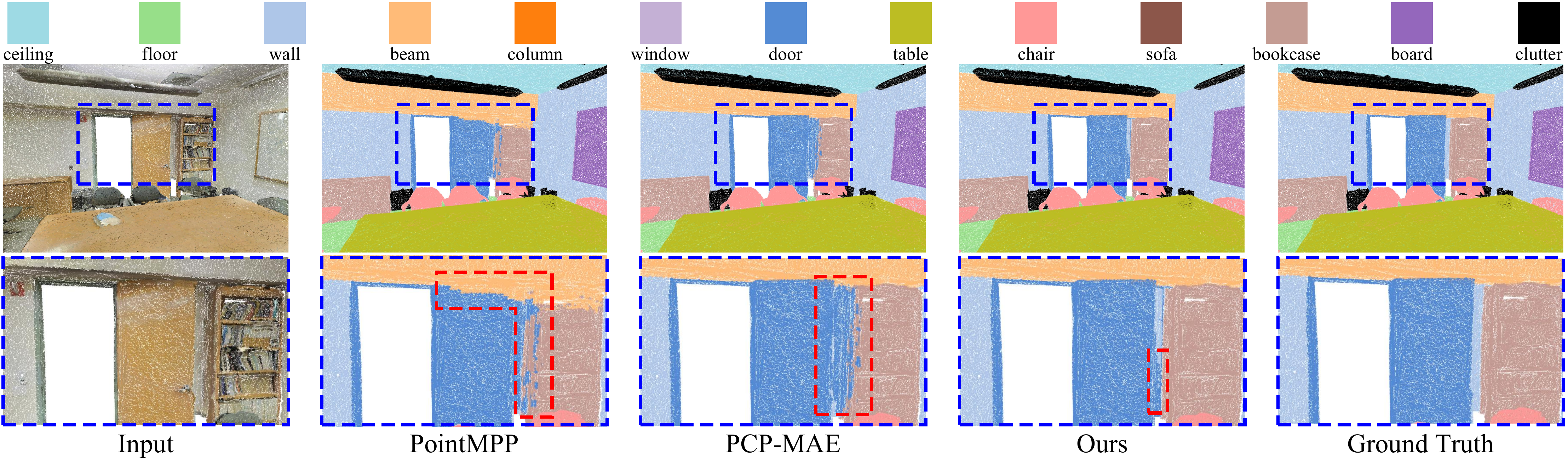}
  \vspace{-6mm}
  \caption{Qualitative comparison on S3DIS semantic segmentation. The first column shows the original point cloud input, followed by columns 2-4, which display the segmentation results of PointMPP, PCP-MAE, and our method. The fifth column shows the ground truth. Regions with poor segmentation results are marked with red dashed boxes.}
  \vspace{-2mm}
  \label{f5}  
\end{figure*}
\begin{figure}[htbp]
  \centering
  \includegraphics[width=\linewidth]{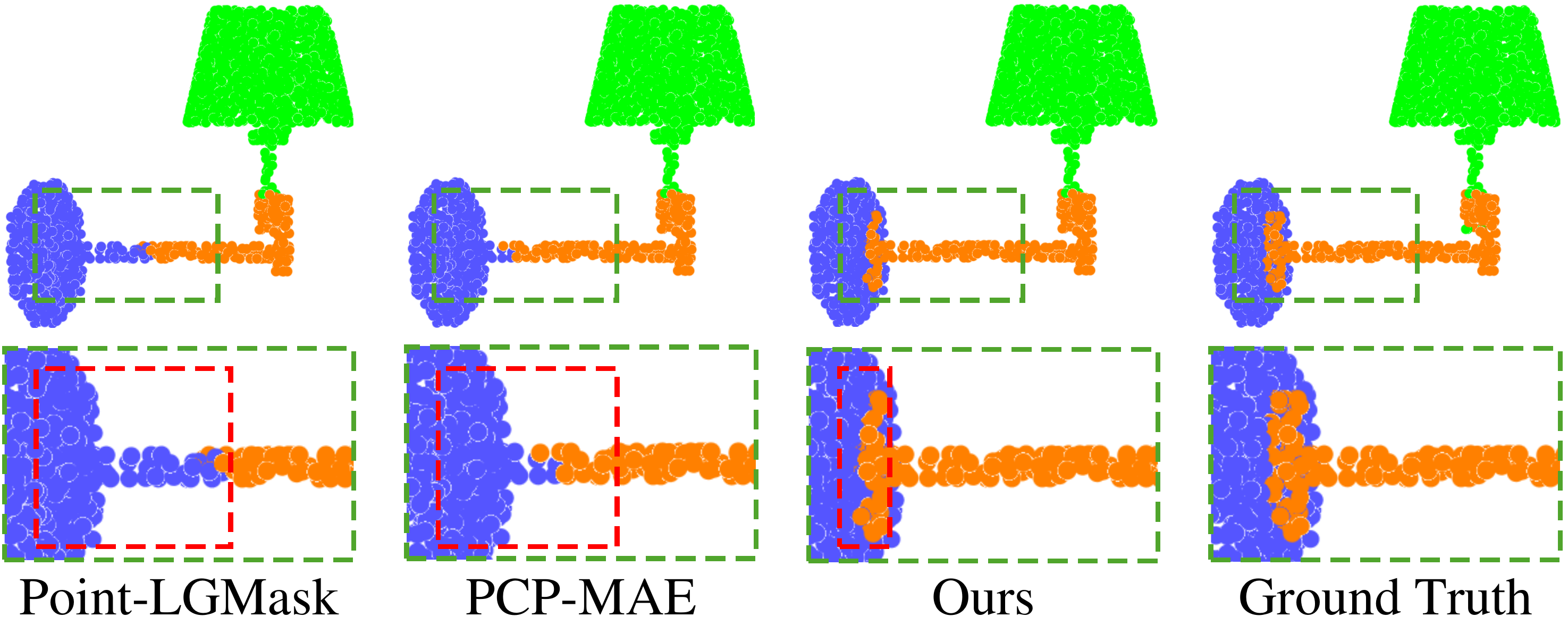}
  \vspace{-5mm}
  \caption{Qualitative comparison of different algorithms on ShapeNetPart, with differences highlighted in red boxes.}
  \label{f4}
  \vspace{-0.5cm}
\end{figure}

\noindent \textbf{\textit{Semantic segmentation}.} We test the generalization of our method on the large-scale S3DIS dataset for semantic segmentation evaluations. The S3DIS dataset consists of 271 scenes from 6 indoor regions, with each point labeled into 13 categories. Following the common practice, we use Area 5 for testing and the remaining areas for training. The experimental results are shown in \cref{t3}. Compared to other methods, \ourmethod achieves the best performance in terms of mIoU and mAcc metrics. Meanwhile, we provide a qualitative comparison of semantic segmentation results on the S3DIS dataset. As illustrated in \cref{f5}, the predictions of our algorithm are closer to the ground truth and less incorrectly segmented result than PointMPP and PCP-MAE.

\noindent \textbf{\textit{Part segmentation}.} We evaluate part segmentation performance of \ourmethod on ShapeNetPart dataset, which consists of 16,881 samples across 16 categories. We partition the point cloud into 128 patches using our geometry-aware partitioning strategy. We use a trilinear interpolation algorithm \cite{44-pointnet++} to up-sample the encoded features to all input points. The experimental results are shown in \cref{t3}. Our method achieves better results in terms of mIoU$_C$ and mIoU$_I$. In our approach, the features are extracted from geometrically homogeneous patches, effectively capturing the distinctive properties of the local geometry. It is beneficial for recovering the features of the original input points. In \cref{f4}, we visualize the segmentation results of different methods. \ourmethod exhibits minimal segmentation errors.

\noindent \textbf{\textit{Point cloud completion}.} Finally, we further validate the effectiveness of \ourmethod on the ShapeNet55 dataset for the point cloud completion task. Following Pointr~\cite{46-pointr}, we randomly remove 25\% to 75\% of points, creating varying levels of incompleteness. The remaining points are then sampled to 2048 points, which are used as the input. We use the original object containing 8192 points as the ground truth. During testing, we set the number of deleted points to 2048, 4096, or 6144. These correspond to three difficulty levels: simple, moderate, and hard. We use the $\ell_2$ Chamfer Distance (CD) and F1-Score\cite{47-F1Score} as the evaluation metrics.

\begin{figure}[htbp]
  \centering
  \includegraphics[width=\linewidth]{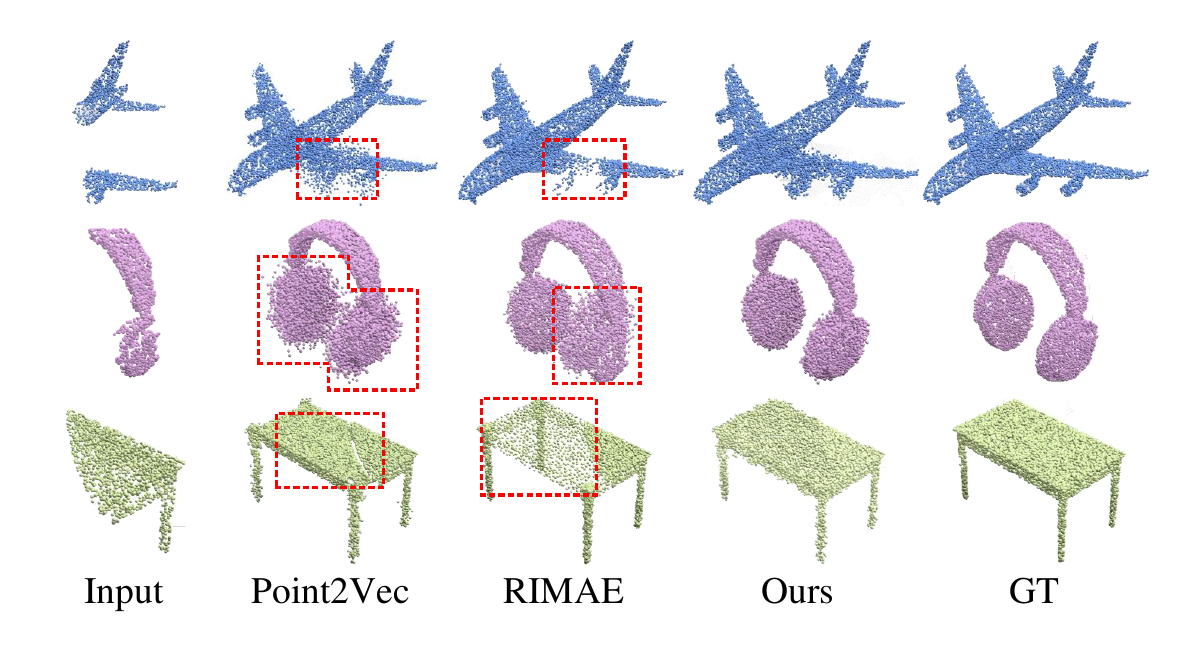}
  \vspace{-0.6cm}
  \caption{Qualitative comparison of different algorithms on the point cloud completion task.}
  \label{f6}
  \vspace{-0.58cm}
\end{figure}

In \cref{t4}, we compare \ourmethod with other self-supervised algorithms whose pretext tasks focus on regressing the mask features. It can be seen that \ourmethod achieves the lowest CD in both moderate and hard settings. In terms of F1-Score, \ourmethod also achieves the best performance, which further indicates that the features extracted by our method are more robust to geometric reconstruction. We also provide a qualitative comparison of the point cloud completion results between the proposed algorithm and other methods in the hard setting. As shown in \cref{f6}, it can be observed that almost all geometric details are lost for the input target. While the results reconstructed by other algorithms remain unsatisfactory, \ourmethod achieves significantly higher fidelity in the reconstructed outputs.

\subsection{Ablation Studies}

\noindent\textbf{\textit{Effects of different components}.} To further validate the effectiveness of the main module design, we conduct extensive experiments on the ModelNet40 dataset. We adopt the Point2Vec\cite{12-Point2vec} method as our baseline model. It employs a teacher-student framework, where the student network learns to regress the feature output from the teacher network. The experimental results are shown in \cref{t5}. It can be observed that the performance of the baseline algorithm is not ideal since regression is a challenging task. For Model B, directly constructing the codebook to perform the MPM task via online k-means did not lead to an improvement. Since the codebook at this stage contains many dead vectors. As observed in Model C, the codebook maintenance mechanism addresses this issue, enabling efficient feature alignment on the codebook for the masked point cloud modeling task. According to Model D, the performance is enhanced when the geometric-aware partitioning strategy replaces the traditional KNN in Model A. Furthermore, the performance reaches its peak when the geometry-aware partitioning strategy is applied to Model C, as demonstrated in Model E. At this point, the geometry represented by the codebook vectors becomes more distinct, enabling them to better serve as cluster centers.

\begin{table}[t]
\vspace{-2mm}
\resizebox{\columnwidth}{!}{
\begin{tabular}{cccccccc}
\toprule
\multirow{2}{*}{Models} & \multirow{2}{*}{Baseline} & \multicolumn{2}{c}{Codebook Update} & \multirow{2}{*}{GAP} & \multirow{2}{*}{\texttt{-}/\texttt{+} Vote} \\
\cmidrule(lr){3-4} 
 &  & \begin{tabular}[c]{@{}c@{}}~~OKC\end{tabular} & CMM &  &  &  \\
 \midrule
A & \checkmark &  &  &  & 92.5 / 93.1 \\
B & \checkmark & ~~\checkmark &  &  & 92.1 / 92.5 \\
C & \checkmark & ~~\checkmark & \checkmark &  & 93.7 / 94.2 \\
\midrule
D & \checkmark &  &  & \checkmark &  92.8 / 93.4 \\
\rowcolor{gray!20} E & \checkmark & ~~\checkmark & \checkmark & \checkmark & \textbf{94.3} / \textbf{94.6} \\
\bottomrule
\end{tabular}
}
\caption{Ablation study of the main modules in the proposed algorithm on ModelNet40. OKC: Online K-means Clustering. CMM: Codebook Maintenance Mechanism. GAP: Geometry-aware Partition Strategy. \texttt{-}/\texttt{+} Vote: without or with the voting strategy.}
\vspace{-0.1cm}
\label{t5}
\end{table}

\begin{table}[t]
\resizebox{\columnwidth}{!}{%
\fontsize{7}{8}\selectfont 
\begin{tabular}{cccccc}
\toprule
\multirow{2}{*}{Group Strategy} & \multicolumn{3}{c}{Mask Ratio} & \multirow{2}{*}{w/o   Vote} & \multirow{2}{*}{w/   Vote} \\
\cmidrule(lr){2-4} 
 & 0.7 & 0.8 & 0.9 &  &  \\
\midrule
KNN & \checkmark &  &  & 93.2 & 93.8 \\
KNN &  & \checkmark &  & 93.7 & 94.2 \\
KNN &  &  & \checkmark & 93.4 & 94.0 \\
GAP & \checkmark &  &  & 93.9 & 94.4 \\
\rowcolor{gray!20}GAP &  & \checkmark &  & \textbf{94.3} & \textbf{94.6} \\
GAP &  &  & \checkmark & 93.6 & 94.2 \\
\bottomrule
\end{tabular}%
}
\caption{Evaluation of the geometric-aware partitioning strategy under different mask ratios.}
\vspace{-0.3cm}
\label{t6}
\end{table}

\noindent\textbf{\textit{Effect of different grouping strategies}.} We validate the impact of different grouping methods for constructing point cloud patches on ModelNet40. Specifically, we compare our method with conventional KNN algorithms under different masking ratios, as shown in \cref{t6}. It can be seen that the performance of the KNN is not ideal under different masking ratios. It leads to information leakage among the obtained point cloud patches, which simplifies the pretext task and ultimately limits the model performance. Benefiting from geometric consistency, the proposed geometry-aware partitioning strategy improves accuracy compared to KNN. Also, appropriately increasing the masking rate can enhance the overall difficulty of the self-supervised pretext task, thus further improving the model performance.

\noindent\textbf{\textit{Effect of different codebook construction formats}.} We present the impact of different codebook construction formats using the ModelNet40 dataset. Considering that codebook vectors should be as active as possible, we adopt a queue-based method for codebook construction, following the approach of MoCo~\cite{50-MOCO}. Inspired by SwAV~\cite{29-SWAV}, we also compare with an alternative implementation using the Sinkhorn-Knopp algorithm~\cite{37-Sinkhorn}. As shown in \cref{t7}, the queue-based codebook ensures all vectors remain active by continuously updating with the latest batch data. However, its vectors do not serve as true cluster centers. Although the Sinkhorn-Knopp algorithm mitigates this by enforcing uniform assignments to $K$ clusters, this assumption is often inconsistent with real-world data distributions, leading to suboptimal performance. The codebook constructed using online k-means can solve this problem. However, the performance does not improve if we simply replace it without further optimization. Because there are many dead vectors in the codebook that have not been updated. After adding meaningful perturbations to the vectors in the codebook as proposed in our design, the model achieves its best performance. It achieves much higher codebook utilization (see in \cref{f3}(c)) which ensures the accuracy of the feature alignment over the codebook.

\begin{table}[t]
\small
\vspace{-2mm}
\resizebox{\columnwidth}{!}{%
\begin{tabular}{lccc}
\toprule
\multirow{1}{*}{Codebook Format} & \multicolumn{1}{c}{Perturbed mode} & \multirow{1}{*}{w/o   Vote} & \multirow{1}{*}{w/   Vote} \\
\midrule
Queue-based & None &  93.7 & 94.1 \\
Sinkhorn-based & None & 93.1 & 93.4 \\
Online k-means & None &  92.7 & 93.2 \\
Online k-means & Random &  93.4 & 93.8 \\
\rowcolor{gray!20}Online k-means & Meaningful & \textbf{94.3} & \textbf{94.6} \\
\bottomrule
\end{tabular}%
}
\caption{Evaluation of the codebook construction format under different perturbed modes.}
\vspace{-0.3cm}
\label{t7}
\end{table}

\section{Conclusion}

In this paper, we propose a novel online codebook-guided teacher-student framework that adopts a clustering-based paradigm to for the MPM task. Unlike traditional regression methods, our method focuses on identifying the cluster centers to which the masked features belong, enabling the model to learn more generalized features. Moreover, by incorporating a geometry-aware partitioning strategy, we ensure geometric consistency within local point cloud block, which enhances the quality of feature learning. To further improves the efficiency of masked feature alignment, we introduce a codebook maintenance mechanism that promotes the convergence of the codebook vectors towards the optimal cluster centroid. Extensive experiments on various downstream tasks validate the effectiveness of the proposed method. Our proposed pre-training framework provides a novel direction to advance point cloud processing.

\section{Acknowledgement} 
This work was supported in part by the National Natural Science Foundation of China under Grant 62271237, Grant U24A20220, and Grant 62132006, in part by the Science Foundation of the Jiangxi Province of China under Grant 20242BAB26014 and Grant 20223AEI91002, and in part by the PNRR FAIR – Future AI Research (PE00000013), funded by NextGeneration EU.

{
    \small
    \bibliographystyle{ieeenat_fullname}
    \bibliography{References}
}

\end{document}